\documentclass[11pt,a4paper]{article}
\usepackage[hyperref]{acl2021}
\usepackage{times}
\usepackage{latexsym}
\usepackage{flushend}

\usepackage{microtype}

\aclfinalcopy 

\usepackage{graphicx}
\usepackage[inline]{enumitem}
\usepackage{multirow}
\usepackage{booktabs}

\newcommand{\ourkb}{\textsc{Ascent}}

\newcommand{\demourl}{https://ascent.mpi-inf.mpg.de}
\newcommand{\youtubeurl}{https://youtu.be/qMkJXqu_Yd4}

\newcommand{\tuple}[1]{\textit{$\langle$#1$\rangle$}}
\newcommand{\term}[1]{\textit{#1}}

\renewcommand{\paragraph}[1]{\smallskip\noindent\textbf{#1.\mbox{\ \ }}}

\title{
Inside \ourkb:
Exploring a Deep Commonsense Knowledge Base
and its Usage in Question Answering
}

\author{Tuan-Phong Nguyen \\
\And
  Simon Razniewski \\
  Max Planck Institute for Informatics \\
  Saarbr\"ucken, Germany \\
  \texttt{\{tuanphong,srazniew,weikum\}@mpi-inf.mpg.de} \\
\And
  Gerhard Weikum \\
}

\date{}

\begin{document}
\maketitle
\begin{abstract}
\ourkb{} is a fully automated methodology for extracting and consolidating commonsense assertions from web contents \cite{ascent}.
It advances traditional triple-based commonsense knowledge representation by capturing semantic facets like locations and purposes, and composite concepts, i.e., subgroups and related aspects of subjects.
In this demo, we present a web portal that allows users to understand its construction process, explore its content, and observe its impact in the use case of question answering.
The demo website\footnote{\url{\demourl}} and an introductory video\footnote{\url{\youtubeurl}} are both available online.
\end{abstract}

\section{Introduction}

Commonsense knowledge (CSK) is an enduring theme of AI~\cite{mccarthy1960programs} that has been recently revived for the goal of building more robust and reliable applications \cite{monroe2020seeking}. Recent years have witnessed the emerging of large pre-trained language models (LMs), notably BERT~\cite{bert}, GPT~\cite{gpt3} and their variants which significantly boosted the performance of tasks requiring natural language understanding such as question answering and dialogue systems \cite{clark2020f}. Although it has been shown that such LMs implicitly store some commonsense knowledge~\cite{olmpics}, this comes with various caveats, for example regarding degree of truth, or negation, and their commercial development is inherently hampered by their low interpretability and explainability.

Structured knowledge bases (KBs), in contrast, give a great possibility of explaining and interpreting outputs of systems leveraging the resources. There have been great efforts towards building large-scale commonsense knowledge bases (CSKBs), including expert-annotated KBs (e.g., Cyc~\cite{lenat1995cyc}), crowdsourced KBs (e.g., ConceptNet~\cite{conceptnet} and Atomic~\cite{atomic}) and KBs built by automatic acquisition methods such as WebChild~\cite{webchild, webchild-2.0}, TupleKB~\cite{tuplekb}, Quasimodo~\cite{quasimodo} and CSKG \cite{cskg}. Human-created KBs, although possessing high precision, usually suffer from low coverage. On the other hand, automatically-acquired KBs typically have better coverage, but also contain more noise. Nonetheless, despite different construction methods, these KBs are all based on a simple subject-predicate-object model, which has major limitations in validity and expressiveness.

We recently presented \ourkb{}~\cite{ascent}, a methodology for automatically collecting and consolidating commonsense assertions from the general web. To overcome the limitations of prior works, \ourkb{} refines subjects with subgroups (e.g., \term{circus elephant} and \term{domesticated elephant}) and aspects (e.g., \term{elephant tusk} and \term{elephant habitat}), and captures semantic facets of assertions (e.g., \tuple{lawyer, represents, clients, LOCATION: in courts} or \tuple{elephant, uses, its trunk, PURPOSE: to suck up water}).

For a given concept, \ourkb{} searches through the web with pattern-based search queries disambiguated using WordNet~\cite{wordnet} hypernymy. Then, irrelevant documents are filtered out based on similarity comparison against the corresponding Wikipedia articles. We then use a series of judicious dependency-parse-based rules to collect faceted assertions from the retained texts. The semantic facets, which come from prepositional phrases and supporting adverbs are then labeled by a supervised classifier. Finally, assertions are clustered using similarity scores from word2vec~\cite{word2vec} and a fine-tuned RoBERTa~\cite{roberta} model.

We executed the \ourkb{} pipeline for 10,000 prominent concepts (selected based on their respective number of assertions in ConceptNet) as primary subjects. In~\cite{ascent}, we showed that the content of the resulting CSKB (hereinafter referred to as \ourkb{} KB) is a milestone in both salience and recall. As extrinsic evaluation, we conducted a comprehensive evaluation of the contribution of CSK to zero-shot question answering (QA) with pre-trained language models~\cite{petroni2020context, realm}. 

This paper presents a companion web portal of the \ourkb{} KB, which enables the following interactions:
\begin{enumerate}[label=\arabic*.]
    \item Exploration of the construction process of \ourkb{}, by inspecting word sense and Wikipedia disambiguation, web search queries, clustered statements, and source sentences and documents.
    \item Inspection of the resulting KB, starting from subjects, predicates, objects, or examining specific subgroups or aspects.
    \item Observation of the impact of structured knowledge on question answering with pretrained language models, comparing generated answers across various CSKBs and QA settings.
\end{enumerate}
%
The web portal is available at \url{\demourl}, and a screencast demonstrating the system can be found at
\url{\youtubeurl}.

\section{\ourkb}
\label{sec:ascent}

Two major contributions of \ourkb{} are its expressive knowledge model, and its state-of-the-art extraction methodology. Details are in the technical paper~\cite{ascent}. In this section, we revisit the most important points.

\subsection{Knowledge model}

\ourkb{} extends the traditional triple-based data model in existing CSKBs in two ways.

\paragraph{Expressive subjects}
Subjects in existing CSKBs are usually single nouns, which implies two shortcomings:
\begin{enumerate*}[label=(\roman*)]
    \item different meanings for the same word are conflated, and
    \item refinements and variants of word senses are missed out.
\end{enumerate*}
\ourkb{} has addressed this problem with the following means:
\begin{enumerate}[label=\arabic*.]
    \item When searching for source texts, \ourkb{} combines the target subject with an informative hypernym from WordNet to distinguish different senses of the word (e.g., ``bus public transport'' and ``bus network topology'' for the subject \term{bus}).
    \item \ourkb{} refines subjects with multi-word phrases into \emph{subgroups} and \emph{aspects}. For example, subgroups for the subject \term{bus} would be \term{tourist bus} and \term{school bus}, while one of its aspects would be \term{bus driver}.
\end{enumerate}

\paragraph{Semantic facets}
The validity of commonsense assertions is usually non-binary \cite{zhang2017ordinal,chalier2020joint}, and depends on specific temporal and spatial circumstances (e.g., lions live for 10-14 years in the wild but for more than 15 years in captivity). Moreover, CSK triples often benefit from further context regarding causes/effects and instruments (e.g., elephants communicate with each other \textit{by creating sounds}, beer is served \textit{in bars}). In \ourkb's knowledge model, such information is added to SPO triples via semantic facets. \ourkb{} distinguished 8 types of facets: \textit{cause, manner, purpose, transitive-object, degree, location, temporal} and \textit{other-quality}.



\begin{figure*}[t]
    \centering
    \includegraphics[width=.75\textwidth]{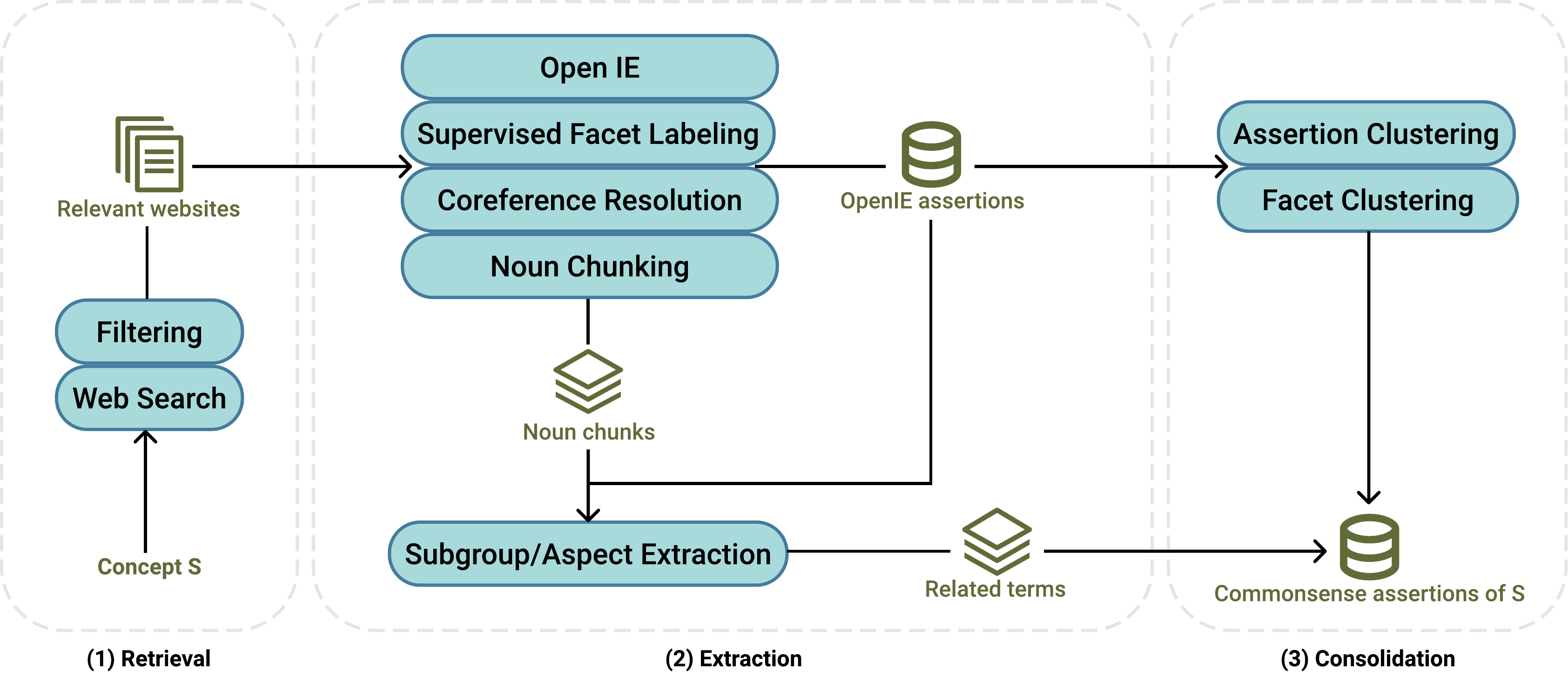}
    \caption{Architecture of the \ourkb{} extraction pipeline \cite{ascent}.}
    \label{fig:architecture}
\end{figure*}

\subsection{Extraction pipeline}
\label{sec:pipeline}
\ourkb{} is a pipeline operating in three phases: \emph{source discovery}, \emph{knowledge extraction} and \emph{knowledge consolidation}. Fig.~\ref{fig:architecture} illustrates the architecture of the pipeline.

\paragraph{Source discovery}
We utilize the Bing Web Search API to obtain documents specific to each subject, with search queries refined by the subject's hypernyms in WordNet. We manually designed query templates for 35 prominent hypernyms (e.g., if subject $s_0$ has hypernym \term{animal.n.01}, we produce the search query ``$s_0$ animal facts'', similarly for the hypernym \term{professional.n.01}, the search query will be ``$s_0$ job descriptions''). We then compute the cosine similarity between the bag-of-words representations of each obtained document and a respective Wikipedia article to determine the relevance of the documents. Low-ranked documents will be omitted in further steps.

\paragraph{Knowledge extraction}
The extractors take in the relevant documents and their outputs include: open information extraction (OIE) tuples, list of subgroups and list of aspects. To obtain \emph{OIE tuples}, we extend the \textsc{StuffIE} approach~\cite{stuffie}, a list of carefully crafted dependency-parse-based rules, to pull out faceted assertions from the texts. Then we classify each facet into one of the eight semantic labels using a fine-tuned RoBERTa model. For \emph{subgroups}, noun phrases whose head word is the target subject are collected as candidates and then are clustered using the hierarchical agglomerative clustering (HAC) algorithm on average word2vec representations. Finally, we collect \emph{aspects} from possessive noun chunks and SPO triples where P is either ``have'', ``contain'', ``be assembled of'' or ``be composed of''.

\paragraph{Knowledge consolidation}
We perform clustering on SPO triples and facet values. As \emph{SPO triples}, we first filter triple-pair candidates with fast word2vec similarity. After that, advanced similarity of triple pairs computed by another fine-tuned RoBERTa model is fed to the HAC algorithm to group the triples into semantically similar clusters. For \emph{facet values}, we group phrases with the same head words together (e.g., ``during evening'' and ``in the evening'').

\subsection{Web portal}
The web portal (\url{\demourl}) is implemented in Python using Django, and hosted on an Nginx web server. The underlying structured CSK is stored in a PostgreSQL database, while for the QA part, statements of all CSKBs are indexed and queried via Apache Solr, for fast text-based querying.
All components are deployed on a virtual machine with access to 4 virtual CPUs and 8 GB of RAM. 


In the demonstration session, we show how users can interact with the portal for exploring the KB (Section~\ref{sec:demo:kb}), understanding the KB construction (Section~\ref{sec:demo:constr}), and observing its utility for question answering (Section~\ref{sec:demo:qa}).

\section{Commonsense QA setups}
\label{sec:qa}

One common extrinsic use case of KBs is question answering. Recently, it was observed that priming language models (LMs) with relevant context can considerably benefit their performance in QA-like tasks~\cite{petroni2020context, realm}. In~\cite{ascent}, to evaluate the contribution of structured CSK to QA, we conducted a comprehensive evaluation consisting of four different setups, all based on the above idea.
\begin{enumerate}[label=\arabic*.]
    \item In \emph{masked prediction} (MP), LMs are asked to predict single masked tokens in generic sentences.
    \item In \emph{free generation} (FG), LMs arbitrarily generate answer sentences to given questions.
    \item \emph{Guided generation} (GG) extends free generation by answer prefixes that prevent the LMs from evading answering.
    \item \emph{Span prediction} (SP) is the task of locating the answer of a question in provided context.
\end{enumerate}

Examples of the QA setups can be seen in Table~\ref{tab:qa-settings}. Generally, given a question, our system will retrieve from CSKBs assertions relevant to it, and then use the assertions as additional context to guide the LMs. In the \ourkb{} demonstrator, we provide a web interface for experimenting with all of those QA setups with context retrieved from several popular CSKBs.

\begin{table}[t]
\centering
\scriptsize
\begin{tabular}{cp{.5\columnwidth}p{.3\columnwidth}}
\toprule
\textbf{Setup} & \textbf{Input} & \textbf{Sample output} \\
\midrule
\multirow{3}{*}{MP} & Elephants eat {[}MASK{]}. {[}SEP{]} Elephants eat roots, grasses, fruit, and bark, and they eat a lot of these things. & everything (15.52\%), trees (15.32\%), plants (11.26\%) \\
\hline
\multirow{4}{*}{FG} & C: Elephants eat roots, grasses, fruit, and bark, and they eat... & They eat a lot of grasses, fruits, and... \\
 & Q: What do elephants eat? & \\
 & A: & \\
\hline
\multirow{4}{*}{GG} & C: Elephants eat roots, grasses, fruit, and bark, and they eat... & Elephants eat a lot of things. \\
 & Q: What do elephants eat? & \\
 & A: Elephants eat & \\
\hline
\multirow{3}{*}{SP} & question=``What do elephants eat?'' & start=14, end=46, \\
 & context=``Elephants eat roots, grasses, fruit, and bark, and they eat...'' & answer=``roots, grasses, fruit, and bark'' \\
\bottomrule
\end{tabular}
\caption{Examples of QA setups \cite{ascent}.}
\label{tab:qa-settings}
\end{table}

\section{Demonstration experience}\label{sec:demonstration}

In the demonstration session, attendees will experience three main functionalities of our demonstration system.

\subsection{Exploring the \ourkb{} KB}
\label{sec:demo:kb}

\begin{figure*}[t]
    \centering
    \includegraphics[width=.85\textwidth]{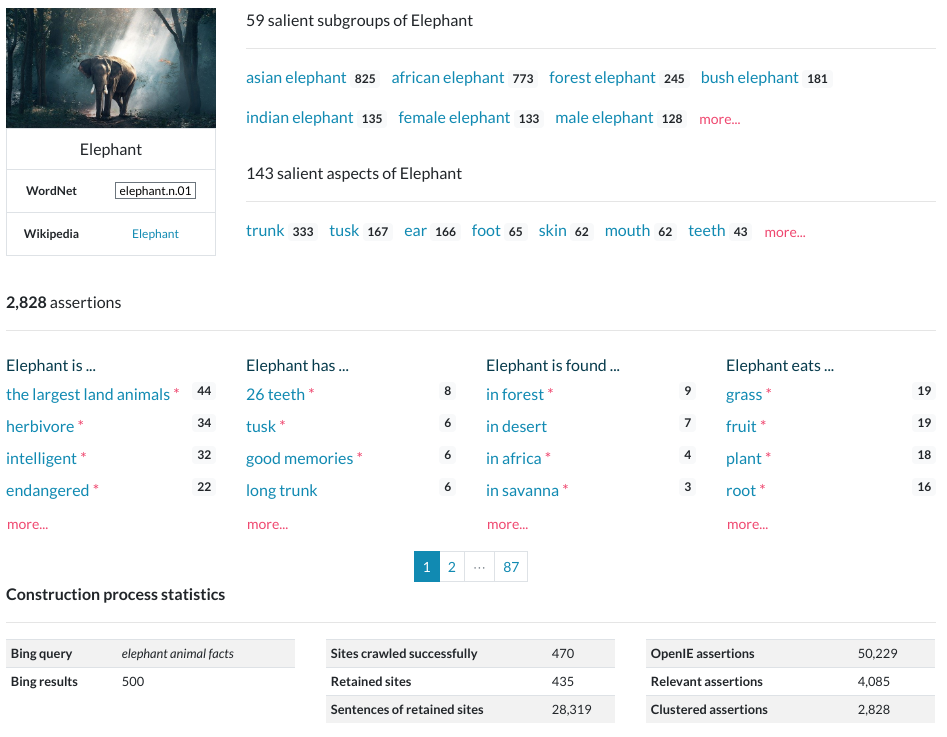}
    \caption{Example of \ourkb{}'s page for the concept \term{elephant}.}
    \label{fig:concept-page}
\end{figure*}

\paragraph{Concept page} Suppose a user wants to know which knowledge \ourkb{} stores for \textit{elephants}. They can enter the concept into the search field in the top right of the start page, and select the first result from the autocompletion list, or press enter, to arrive at the intended concept. The resulting website (see Fig.~\ref{fig:concept-page}) is divided into three main areas. 

At the top left, they can inspect an image from \url{https://pixabay.com}, the WordNet synset used for disambiguation, the Wikipedia page used for result filtering, and a list of alternative lemmas, if existing.

At the top right, users can see subgroups and related aspects, which in our knowledge representation model, can carry their own statements. This way, they can learn that the most salient aspects of \textit{elephants} are their \textit{trunks, tusks} and \textit{ears}, or that \textit{elephant trunks have more than 40,000 muscles}.

    The body of the page, presents the assertions, organized into groups of same-predicate assertions. In each group, assertions are sorted by their frequency displayed beside their objects. For example, the most commonly mentioned \textit{foods} of \textit{elephants} are \textit{grasses, fruits,} and \textit{plants}.
    Many assertions come with a red asterisk. This indicates that the assertion comes with semantic facets. When clicking on an assertion, it will show a small box displaying an SVG-based visualisation of the assertion in which we illustrate all elements of the assertion: its subject, predicate, object, facet labels and values, frequency of the assertion as well as frequency of each facet. For example, one can see that the purpose of \textit{elephants using their trunks} is \textit{to suck up water}.



\paragraph{Searching and downloading assertions}
Alternatively to exploring statements starting from a subject, users can start from a search functionality under the \textit{Browse} menu. This way, they can search, for instance, for all concepts that eat grass (\textit{capybara, zebra, kangaroo, ...}).

The website also provides a JSON-formatted data dump (678MB) of all 8.9 million assertions extracted by the pipeline and their corresponding source sentences and documents. This dataset is also accessible via the HuggingFace Datasets package\footnote{\url{https://huggingface.co/datasets/ascent_kb}}.

\subsection{Inspecting the construction of assertions}
\label{sec:demo:constr}
For many downstream use cases, it is important to know about the provenance of information. 

Users can inspect general properties of the construction process by observing the WordNet lemma and the Wikipedia page used for filtering, as well as inspect specific statistics about the number of retained websites, sentences, and assertions, in a panel at the bottom of subject pages (e.g., 435 websites were retained for \textit{elephant}, from which 50k OpenIE assertions could be extracted).

Furthermore, users can look deeply into the construction process of each assertion on its own dedicated page, which displays the following:
\begin{enumerate}[label=\arabic*.]
    \item \emph{Clustered triples}: These are triples that were grouped together in the knowledge consolidation phase (cf. Section~\ref{sec:pipeline}), where the most frequent triple was selected as cluster representative. For example, for the assertion \tuple{lion, eat, zebra, DEGREE: mostly} (14), the cluster contains: \tuple{lion, eat, zebra} (9), \tuple{lion, prey on, zebra} (2), \tuple{lion, feed on, zebra} (1), \tuple{lion, feed upon, zebra} (1), \tuple{lion, prey upon, zebra} (1). The numbers in parentheses indicate their corresponding frequency.
    \item \emph{Facets}: The assertion's facets are presented in a table whose columns are facet value, facet type and clustered facets. The frequency of each clustered facet is also indicated.
    \item \emph{Source sentences and documents}: Finally, we exhibit the sentences from which the assertions were extracted and their parent documents (in the form of URLs). Furthermore, in the extraction phase, we also recorded the position of assertion elements (i.e., subject, predicate, object, facet) in the source sentences. We show that information to users by highlighting each kind of element with a different color in the source sentences.
\end{enumerate}


\subsection{Experimenting with commonsense QA}
\label{sec:demo:qa}
The third functionality experienced in the demo session is the utilization of commonsense knowledge for question answering (QA).

\paragraph{Input}
There are four main parts in the input interface for the QA experiment:
\begin{enumerate}[label=\arabic*.]
    \item \emph{QA setup}: The user chooses one QA setup they want to experiment with. Available are \emph{Masked Prediction}, \emph{Span Prediction} and \emph{Free/Guided Generation}. If \emph{Masked Prediction} is selected, the user can choose how many answers the LM should produce. For the \emph{Generation} settings, users can provide an answer prefix to avoid overly evasive answers.
    \item \emph{Input query}: The user enters the text question as input. The question can be in the form of a masked sentence (in the case of \emph{Masked Prediction}), or a standard natural-language question (in other setups).
    \item \emph{Retrieval options}: The user can select one supported retrieval method and the number of assertions to be retrieved per CSKB for each question.
    \item \emph{Context sources}: The user selects the sources of context (i.e., ``no context'', CSKBs and ``custom context''). If a CSKB is selected, the system will retrieve from that KB assertions relevant to the given input question. If ``custom context'' is selected, user must then enter their own content. The ``no context'' option is available for all setups but \emph{Span Prediction}.
\end{enumerate}


\paragraph{Output}
The QA system presents its output in the form of a table which has three columns: \textit{Source}, \textit{Answer(s)} and \textit{Context}. For \emph{Masked Prediction} and \emph{Span Prediction}, answers are printed with their respective confidence scores, meanwhile for \emph{Free/Guided Generation}, only answers are printed. For \emph{Span Prediction} in which answers come directly from given contexts, we also highlight the answers in the contexts.

\begin{figure*}[t]
    \centering
    \includegraphics[width=.85\textwidth]{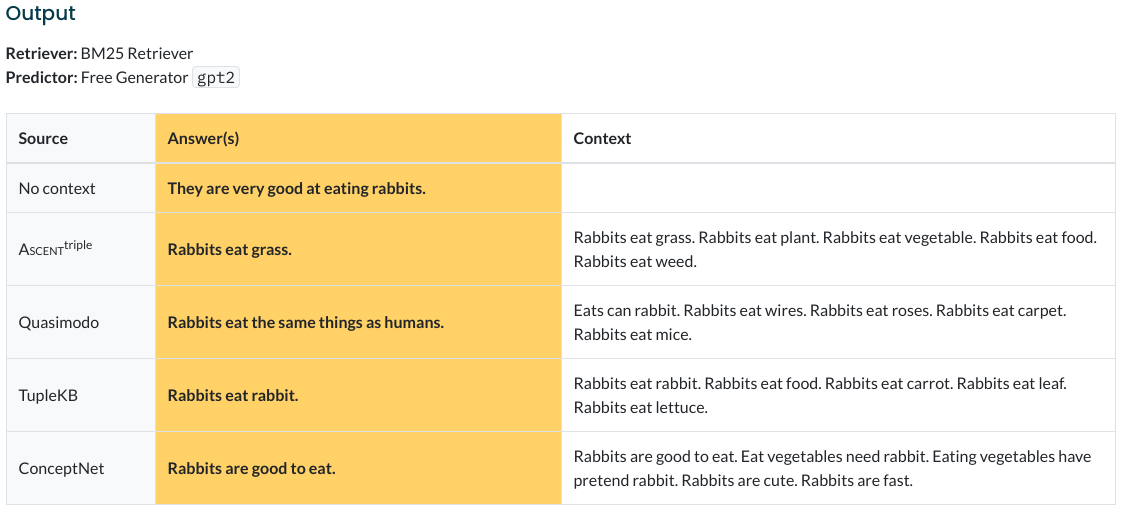}
    \caption{\emph{Free Generation} output for question: \textit{``What do rabbits eat?''}.}
    \label{fig:qa-output-generative}
\end{figure*}

An example of the QA demo's output for the question \textit{``What do rabbits eat?''} under the \emph{Free Generation} setting can be seen in Fig.~\ref{fig:qa-output-generative}. One can observe that language models' predictions are heavily influenced by given contexts. Without context, GPT-2 is only able to generate an evasive answer. When being given context, it tends to re-generate the first sentence in the context first, (e.g., see the answers aligning with \ourkb{}, TupleKB and ConceptNet in Fig.~\ref{fig:qa-output-generative}). For the context retrieved from Quasimodo, GPT-2 is able to overlook the erroneous first sentence, however its generated answer is rather elusive despite the fact that subsequent statements in the context all contain direct answers to the question.

The question \textit{``Bartenders work in [MASK].''} under the \emph{Masked Prediction} setting is another example for the influence of context on LMs' output. Since \term{bartender} is a subject well covered by the \ourkb{} KB, the assertions pulled out are all relevant (i.e., \textit{Bartenders work in bar. Bartenders work in restaurant}\dots) which help guide the LM to a good answer (\textit{bar}). Meanwhile, because this subject is not present in TupleKB, its retrieved statements are rather unrelated (\textit{Work capitals have firm. Work experiences include statement}\dots). Given that, the top-1 prediction for this KB was \textit{tandem} which is obviously an evasive answer.



\section{Related work}\label{sec:related-work}

\paragraph{CSKB construction}
Cyc~\cite{lenat1995cyc} is the first attempt to build a large-scale commonsense knowledge base. Since then, there have been a number of other CSKB construction projects, notably ConceptNet~\cite{conceptnet}, WebChild~\cite{webchild, webchild-2.0}, TupleKB~\cite{tuplekb}, and more recently Quasimodo~\cite{quasimodo}, Dice~\cite{chalier2020joint}, Atomic~\cite{atomic}, and CSKG~\cite{cskg}. The early approach to building a CSKB is based on human annotation (e.g., Cyc with expert annotation and ConceptNet with crowdsourcing annotation). Later projects tend to use automated methods based on open information extraction to collect CSK from texts (e.g., WebChild, TupleKB and Quasimodo). Lately, CSKG is an attempt to combine various commonsense knowledge resources into a single KB. The common thread of these CSKB is that they are all based on SPO triples as knowledge representation, which has shortcomings~\cite{ascent}. 
\ourkb{} is the first attempt to build a large-scale CSKB with assertions equipped with semantic facets built upon the ideas of semantic role labeling~\cite{palmer2010semantic}.

\paragraph{KB visualization}
Most CSKBs share their content via CSV files. Some, like ConceptNet\footnote{\url{http://conceptnet.io}}, WebChild\footnote{\url{https://gate.d5.mpi-inf.mpg.de/webchild}}, Atomic\footnote{\url{https://mosaickg.apps.allenai.org/kg-atomic2020}} and Quasimodo\footnote{\url{https://quasimodo.r2.enst.fr}}, have a web portal to visualise their assertions.
The most common way for CSKB visualisation is to use a single page for each subject and group assertions by predicate (e.g., in ConceptNet and WebChild). Quasimodo, on the other hand, implements a simple search interface to filter assertions and presents assertions in a tabular way~\cite{romero2020inside}. The \ourkb{} demo has both functionalities: exhibiting assertions of each concept in a separated page, and supporting assertion filtering. Our demo also uses an SVG-based visualisation of assertions with semantic facets, which are a distinctive feature of the \ourkb{} knowledge model.

\paragraph{Context in LM-based question answering}
Priming large pretrained LMs with context in QA-like tasks is a relatively new line of research \cite{petroni2020context,realm}. In our original paper, we made the first attempt to evaluate the contribution of CSKB assertions to QA via four different setups based on that idea. While others use commonsense knowledge for (re-)training language models~\cite{hwang2020comet,ilievski2021dimensions,ma2020knowledge,mitra2020additional}, to the best of our knowledge, our demo system is the first to visualize the effect of priming vanilla language models, i.e., without task-specific retraining.

\section{Conclusion}\label{sec:conclusion}
We presented a web portal for a state-of-the-art commonsense knowledge base---the \ourkb{} KB. It allows users to fully explore and search the CSKB, inspect the construction process of each assertion, and observe the impact of structured CSKBs on different QA tasks. We hope that the portal enables interesting interactions with the \ourkb{} methodology, and that the QA demo allows researchers to explore the potentials of combining structured data with pre-trained language models.


\bibliographystyle{acl_natbib}
\typeout{}
\bibliography{refs}


\end{document}